\documentclass{article}

\usepackage{microtype}
\usepackage{graphicx}
\usepackage{subfigure}
\usepackage{booktabs} 
\newcommand\given[1][]{\:#1\vert\:}
\usepackage{amssymb}
\usepackage{amsmath}
\usepackage{epstopdf}

\DeclareGraphicsExtensions{.eps}

\usepackage{setspace}

\usepackage{hyperref}

\usepackage[accepted]{icml2020}

\icmltitlerunning{Towards Scheduling Federated Deep Learning using Meta-Gradients for Inter-Hospital Learning}

\begin{document}

\twocolumn[
\icmltitle{Towards Scheduling Federated Deep Learning using Meta-Gradients for Inter-Hospital Learning}



\icmlsetsymbol{equal}{*}

\begin{icmlauthorlist}
\icmlauthor{Rasheed el-Bouri}{UoO}
\icmlauthor{Tingting Zhu*}{UoO}
\icmlauthor{David A. Clifton*}{UoO}

\end{icmlauthorlist}
\icmlaffiliation{UoO}{University of Oxford}


\icmlcorrespondingauthor{Rasheed el-Bouri}{rasheed.el-bouri@eng.ox.ac.uk}


\icmlkeywords{Machine Learning, ICML}

\vskip 0.3in
]



\printAffiliationsAndNotice{\icmlEqualContribution} 

\begin{abstract}
    Given the abundance and ease of access of personal data today, individual privacy has become of paramount importance, particularly in the healthcare domain. In this work, we aim to utilise patient data extracted from multiple hospital data centres to train a machine learning model without sacrificing patient privacy. We develop a scheduling algorithm in conjunction with a student-teacher algorithm that is deployed in a federated manner. This allows a central model to learn from batches of data at each federal node. The teacher acts between data centres to update the main task (student) algorithm using the data that is stored in the various data centres. We show that the scheduler, trained using meta-gradients, can effectively organise training and as a result train a machine learning model on a diverse dataset without needing explicit access to the patient data. We achieve state-of-the-art performance and show how our method overcomes some of the problems faced in the federated learning such as node poisoning. We further show how the scheduler can be used as a mechanism for transfer learning, allowing different teachers to work together in training a student for state-of-the-art performance.
\end{abstract}

\section{Introduction}
Federated learning is a field that has emerged recently due to the abundance of data available today and the risks that this poses to individuals. Privacy (particularly of personal information) is of great importance and should be protected by researchers working in machine learning. In parallel, the emergence of electronic health records (EHRs) has allowed the digitisation of much personal information pertaining to the health conditions of individuals. EHRs are often used in many machine learning research projects \cite{shailaja2018machine}. This often involves the transfer and storage of very sensitive information which increases the risk of data leakage. As a result, federated learning can minimise this risk by utilising the data at it's source rather than transferring it to the researchers servers to be processed.
\par Machine learning researchers working with EHR data will be all too familiar with the difficulty of gaining access to this information in the first place. It can be a very lengthy and exhausting process (for good reason) to gain access and utilise the EHRs of one healthcare institution let alone accessing the datasets of many. Federated learning offers an alternative in that it allows the data of the patients to be utilised while reducing the risk of their privacy being compromised.

\par Federated learning is not without its own limitations however. The different datasets that are stored in the different nodes may have different underlying distributions due to their data collection processes which can make machine learning across multiple domains difficult. There is also the possibility of data at each node being corrupted either maliciously or accidentally, leading to data that is undesirable to use for training. These issues can all lead to difficulties in training and convergence of the overall model being trained. 

\par To overcome these limitations, in this work we propose the following setup. Firstly, We use federated learning to i) protect the privacy of patients by minimising the movement of their data and ii) improve the quality of our machine learning model by utilising a diverse dataset sourced from different hospitals. As we are `blind' to the data at the nodes, we propose the use of a student-teacher network setup. The teacher (a reinforcement learning agent) will have access to the local servers and be able to select the appropriate data for the `student' (our model) to be trained on at that time.  The `scheduler' will be responsible for directing the teacher to a given data centre to select data for training on.
\par To summarise: (\textit{The Student}:) - the machine learning model we are training. (\textit{The Teacher}:) - a reinforcement learning agent that selects data from a data centre based on the state of the student. This essentially defines a curriculum at each training step for the student. (\textit{The Scheduler}:) - directs the teacher to the appropriate data centre for training. This is also based on the state of the student at each training iteration.

Section \ref{Section:related_work} discusses the related work that has been carried out and Section \ref{Section:Methodology} provides a detailed description of how our algorithm works. Section \ref{Section:Datasets} details the datasets we benchmark our method against. We then present the results in Section \ref{Section:results} and discuss their significance and interesting behaviours of our model in Section \ref{section:defense}.

\section{Related Work}
\label{Section:related_work}

Federated learning has been used by researchers to exploit larger pools of data for training \cite{bonawitz2019towards}, preserve data privacy \cite{xu2019hybridalpha} and distributing computational resource requirements \cite{yuan2020federated}. Federated learning has also been used for healthcare applications to simultaneously utilise multiple datasets to train a model on patient data.

In this work we create a model that learns in a federal fashion through the interaction of a scheduler that is trained using meta-gradients and a student-teacher algorithm that is trained using reinforcement learning. 
\par Meta-learning has been used effectively in \cite{such2019generative} where the loss of a student model on a validation set was used as a signal to update the weights of a generative model. This work demonstrated the rapid and effective training of methods that exploit meta-gradients. Meta-learning was also used in \cite{zahavy2020self}, where the meta-gradients are used to tune the parameters of an actor-critic algorithm. As a result of the efficacy of this method in these domains we choose to use meta-gradients in order to schedule which data centre the gradients to update are student model will come from.
\par The meta-learned scheduler chooses a node representing a data centre where a student-teacher algorithm is used to sample data. Student-teacher algorithms have been used in multiple works, with the general premise that one algorithm (teacher) is trained to train another (student) \cite{fan2018learning, liu2017iterative}. These methods have also been used with a curriculum \cite{bengio2009curriculum}, where the curriculum is either pre-defined and exploited by the teacher \cite{el2020student} or implicitly learned by the teacher during training \cite{graves2017automated}.

Federated learning is a method of training a model (in our case a deep neural network) by using data from multiple centres, without having central access to each of them \cite{mcmahan2017communication}. Local models at each of the data centres are iteratively updated and aggregated to form a global model. At each round of iteration, a central coordinator samples a subset, $m$, of local models, $S_m$, and sends them the current global model $G^t$. Each member of $S_m$ then updates this global model using their local data to create an updated model $L^{t+1}$. These models are then aggregated and are sent back to update the global model as:
\begin{equation}
    G^{t+1} =  G^t + \frac{\eta}{n}\sum^{m}_{i=1}\left(L^{t+1}_{i} - G^t\right)
\end{equation}
where $n$ is the number of \textit{nodes} (i.e., data centres) and $\eta$ acts as a learning rate for replacing the global model with the aggregate of the local models. While this has been shown to work in many cases, \cite{wang2020federated} make the argument that there is inherent difficulty in updating neural networks in this manner. They argue that the permutation invariance of the summation operand renders averaging in the parameter space a naive approach. For meaningful averaging to be done, the permutation must first be undone. 

\subsection{Compromising Federated Learning}
One of the vulnerabilities of federated learning is that nodes being compromised can significantly affect the training of the global model \cite{bhagoji2018model}. Attacks of these sort can either `poison' the data found at one of the nodes (known as an \textit{adversarial} attack) or bias the model that is trained at one of the nodes significantly, leading it to highly skew the aggregation step \cite{tolpegin2020data}. There is also the possibility of the attack being a \textit{single-shot attack} or a \textit{repeated attack} \cite{fang2020local}. In the single shot case, only one of the nodes is compromised whereas in the repeated case, multiple nodes can be compromised at any given time. Many works have been produced in discussing how federated learning can be compromised by introducing a backdoor into the training process \cite{gu2017badnets, bagdasaryan2020backdoor}. A backdoor is an attack that causes a classifier to produce unexpected behaviour if a specific trigger is added to the input space. An example is a sticker being added to an image and associating this with the incorrect label \cite{gu2017badnets}. 
\par Defences against these attacks have been developed with some authors using pruning of redundant neurons for the core classification task \cite{liu2018fine}, using outlier detection to detect potential triggers \cite{wang2019neural}, and re-training and preprocessing inputs \cite{liu2017neural}.
 
In this work we aim to overcome these limitations and build defence into the training procedure through the use of a student-teacher network that actively selects which data to train on.

\section{Methodology}
\label{Section:Methodology}
Our method is comprised of three agents in the training setup, the student, the teacher and the scheduler.

\subsection{The Overall Setup}

The overall setup of our federated learning training routine is as follows. We have a scheduler that controls which node we will be learning from (this can be one-hot or we can select multiple nodes). The teacher at the node can then select a batch of data according to the state of the student. The student at the node is a copy of the global student. We use the student to forward pass the batch of data selected by the teacher and return the loss. In the one-hot scheduler scenario, we send back the loss to the global student model to update the weights via backpropagation. In the multi-node learning scenario, we aggregate the losses from all nodes selected and feed these back to the global model for updating. 
\subsection{Data Preprocessing}
The first step we must take in order to exploit our teacher setup is to rank our data according to some metric. Using \cite{wang2020comprehensive} as a guide, we choose to use the Mahalanobis distance expressed as:
\begin{equation}
d\left({\mathbf{x}_n}\right) = \left(\left(\mathbf{x_n} - \boldsymbol{\mu} \right)^T \mathbf{S^{-1 }}\left({\mathbf{x_n} - \boldsymbol{\mu}}\right)\right)^\frac{1}{2}
\label{eq:Mahalanobis_distance}
\end{equation}
for medical datasets, and the cosine similarity as our similarity metric for image datasets. 

As the tabular data found in electronic health record systems consist of multiple data types, we encode these using a denoising autoencoder. This trained encoder is distributed to all the nodes so that the data in each node is processed in the same way for consistency.
\subsection{The Teacher}

For the student-teacher interaction we follow the setup in \cite{el2020student}. The task of the teacher is to select a batch of data from the curriculum by selecting the index along the curriculum and the `width' around that index to include in the selection. The following sequential steps are implemented:
\begin{itemize}
    \item The data at each node is organised into $N$ curriculum batches according to some metric $H$. 
    \item The teacher selects one or more batches for feeding into the student.
    \item A pre-trained autoencoder is used to create a latent representation of the batch.
    \item The student is trained on this batch and it's performance on a separate validation set is recorded.
\end{itemize}

In this work we use the Mahalanobis distance for $H$ for medical data, and cosine similarity for image data as summarised in \cite{wang2020comprehensive}.

The teacher is a reinforcement learning agent and therefore is tasked with minimising the Bellman loss function given by:
\begin{equation}
    \label{eq:Bellman_loss}
    \mathcal{L}(\theta_i) = \left(r + \gamma\max_{a'}Q(s',a';\theta^{-}_i) - Q(s,a;\theta_i) \right)^2
\end{equation}
where $r$ is the reward of a state-action $(s,a)$ tuple, $\gamma$ the discount factor, $Q$ is the q-value defining the value of taking an action given a state and $\theta$ and $\theta^{-}$ are the parameters of the prediction and target (the version of the teacher that is held constant for $K$ steps to stabilise training as described in \cite{mnih2015human}) networks respectively.

As we also choose to use an actor-critic setup for the teacher, the action space and Q-function are separately parameterised. This allows a continuous action space and the actor that selects actions is updated using the following loss:
\begin{equation}
\label{eq:DDPG_loss}
    \nabla \theta^{\mu} J \approx \frac{1}{N} \sum_i \nabla_a Q\left(s,a \given \theta^Q\right)\given_{s=s_i, a=\mu\left(s_i\right)} \nabla_{\theta^{\mu}} \mu\left(s\given \theta^{\mu}\right)
\end{equation}
where $Q$ is the Q-function and $\mu$ is the policy.

The teacher can either be pre-trained, or jointly trained with the scheduler. The teacher can also either be trained on the dataset of one node and distributed to the rest or independently trained at each node. The latter is preferable due to the ability of the teacher to adapt to the dataset at hand. However, the former is useful when not all nodes in the federated system have access to computational power. The intuition is that the curriculum strategy learned by the teacher should be general for the task at hand and thereby provide strong performance.

\subsection{The Student}
The input to the teacher is the current state of the student. The student in this work is a feedforward neural network that is tasked with classification. The state of the student is defined as a representation of the weights of the student. Given a matrix of weights, $W^{ij}$, between layers $i$ and $j$ of the network, for each row, $W^{ij}_{n:}$, we take the inner product of the row with a fixed reference vector $a$. From this inner product we extract   $\lvert \langle W_{n:}^{ij}, a \rangle \rvert$ and $\angle\left(W_{n:}^{ij}, a\right)$ for $n = 1,2, \dots , M_i$ where $M_i$ is the number of hidden nodes in layer $i$. These values are concatenated to represent the row and this process is repeated for all rows to build the vector. For more hidden layers the process is repeated until we have one vector representing the network. This provides us with a representative vector, $\mathbf{v} \in \mathbb{R}^{2\left(\sum_{l}^{h} M_l\right)}$, where $h$ is the number of hidden layers. This vector is what is fed to the teacher to understand the state of the student.

\subsection{The Scheduler}
The scheduler is the last of our agents in the training setup. This is also a neural network that takes the student state as input and selects which of the nodes the training data should come from at the current iteration of training. This agent is trained using the meta-gradients generated from validation losses similarly to how they are employed in \cite{such2019generative}. We have an `inner' loop of training whereby the student-teacher interaction takes place. In the `outer' loop, we aggregate the losses on validation sets at each node and use this aggregated loss as the signal to update the weights of our scheduler. 
\begin{figure}[h!]
    \centering
    \includegraphics[scale=0.26]{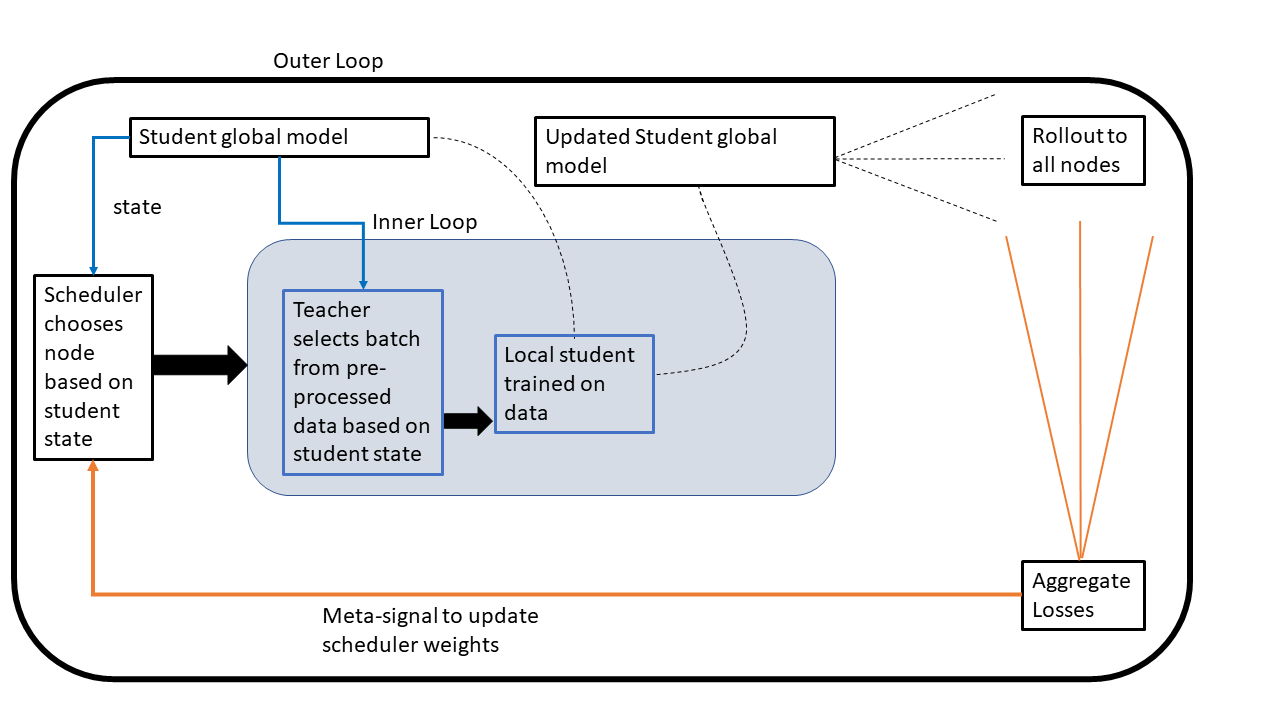}
    \caption{A diagram displaying what happens in the training routine every iteration. The black arrows indicate selection, the blue indicate the extraction of state, the dashed indicate the transfer of a model and the orange indicate the movement of losses.}
    \label{fig:iteration_training}
\end{figure}
Figure \ref{fig:iteration_training} shows diagrammatically the training procedure at every iteration of training. In the outer loop the scheduler selects the node(s) to use for data selection. The global student model is sent to the node and the teacher at the node then selects the data in the inner loop and trains the student network on this. The student is then sent back to the central node and distributed to all nodes. This student is tested on separate validation sets at each node and their losses are aggregated by summing them. They are then used as the loss to update the scheduler. The inner loop loss function is dependent upon the target task with crossentropy used for classification and mean squared error used for regression. The scheduler is updated as:
\begin{equation}
    \label{eq:scheduler_loss}
    \theta_{sc}^{t+1}  = \theta_{sc}^t + \omega \sum_{0 \leq t' \leq t} \alpha^{t-t'} \nabla \left(\mathcal{L}_{inner} \right)
\end{equation}
where $\mathcal{L}_{inner}$ is given by:
\begin{equation}
    \label{eq:scheduler_loss_inner}
    \ell_{inner}\left(T\left(\mathbf{s}_{t}; \theta_{te}\right),\mathbf{y}^*_{t}; \theta_{st}\right) \\
     + \sum_{n=1}^{N-1} \ell_{inner}\left(\mathbf{x}^n_{v}, \mathbf{y}^n_{v}; \theta_{st}\right)
\end{equation}
where $\theta_{sc}$ are scheduler weights, $\omega$ is a learning rate for stochastic gradient descent, $\alpha$ is a momentum hyperparameter, $\ell_{inner}$ is our local task loss function, $T$ is the teacher network taking as input the student state at iteration $t$, $\mathbf{s}_t$, $\mathbf{y^*_t}$ is the ground truth associated with the teacher selection and $\theta_{te}$ and $\theta_{st}$ the teacher and student parameterisations respectively. $\mathbf{x}^n_v$ and $\mathbf{y}^n_v$ are the features and labels of the validation set of node $n$ respectively.

\subsubsection*{Entropy Loss}
As the scheduler is trained using meta-gradients, the model may converge after a set number of iterations. While we would like convergence of the student (i.e., the task solving model), we do not necessarily need the scheduler to converge. In fact using the loss on the validation set as the signal to update the scheduler automatically prevents the scheduler from converging for very long. This is because, should the scheduler converge on selecting a particular data centre, the validation scores on the other data centres will deteriorate thereby increasing the aggregated loss on the validations and providing an update signal to the scheduler. This however was found in practice to require many iterations of training and so in order to encourage the exploration further we add an \textit{entropic loss} term to the scheduler. This takes the form:
\begin{equation}
    \ell_{ent} = \frac{1}{H\left[S\left(s_t;\theta^t_{sc}\right)\right] + \epsilon}
\end{equation}
where $H$ is entropy, $S\left(s_t;\theta^t_{sc}\right)$ is the softmax output of the scheduler and $\epsilon$ is a small positive value (we use $10^{-5}$) to prevent potential division by zero. $\ell_{ent}$ is then added on to the end of the expression shown in Equation \ref{eq:scheduler_loss_inner} to discourage fast convergence of the scheduler.
\par For tabular classifications, we define the student to be a feedforward neural network consisting of 2 hidden layers and 50 nodes in each layer. These are all activated with ReLU activations. For the image recognition tasks we initialise a student that has 4 convolutional layers with 32 filters of size 3x3 in the first two layers and 64 of these in the second two. These are all activated by ReLU and maxpooled and are followed by 3 feedforward layers of size 50 nodes each. 
\par For the results reported in Section \ref{Section:results}, we use a pre-trained teacher (i.e., the teacher has been trained using reinforcement learning on different students for the same classification problem). The teacher has 2 hidden layers and 150 nodes each activated by ReLU apart from the final layer (of size 2) which is activated by a tanh function. For the scheduler we use a feedforward neural network with 2 hidden layers and 100 nodes in each layer activated by ReLU. The output is activated by a softmax of size the number of nodes in the federated system.

\section{Datasets}
\label{Section:Datasets}
In this study we considered the patient data collected in the electronic health records (EHR) of the Other Unknown Hospital (OUH), which the authors are associated with. The data is split randomly in $N$ datasets so that each can act as a separate machine in a federated system. The features include demographic, physiological and medical information (such as age, heart rate upon entry and any medical tests requested by clinical staff who greet the patient). We aim to predict which department in the hospital the patient will consume resource from (i.e., which department in the hospital will ultimately be responsible for treating the patient) rendering this a seven-class classification. In carrying out this classification, this allows hospitals to predict their resource requirements ahead of time and update scheduling and planning accordingly. Only patients who were admitted in an emergency were considered providing a dataset of 14,324 patients. A training set of 60\% of the dataset was used and was balanced, leaving 8,589 patients for training on. The validation set was 20\% of the dataset and testing was also 20\%. These sets are then evenly divided according to the number of nodes in the federated system. The full feature set is included in the supplementary material.

\subsubsection*{eICU}
In order to validate our results on real-world data collected from different hospitals, we introduce the eICU dataset \cite{pollard2018eicu} also hosted on Physionet \cite{goldberger2000physiobank}. The task here is mortality prediction (binary classification) based on features extracted from admission to the ICU as is done in \cite{sheikhalishahi2020benchmarking}. As this dataset contains identifiers for individual hospitals, we are able to create nodes corresponding to each hospital. The features selected are as outlined in the appendix. We choose to learn from the eight hospitals with the largest populations in the dataset leaving us with 8,594 instances. We sample 60\% from each node to keep as the training set, and keep 20\% for the validation set and the final 20\% as the test set. As per usual, the validation set is kept on the local node for performance aggregation during the scheduler training.

\subsubsection*{CIFAR-10}
To test our methodology on the image space we also report results on the CIFAR-10 image recognition dataset \cite{krizhevsky2009learning}. We use 40,000 training examples for the training set and 10,000 each for validation and test set examples. We once again randomly divide the dataset into $N$ datasets to mimic a federated learning system. It should be noted that in Table \ref{table:max_performance_results} the CIFAR-10 dataset has been split into three samples of size 15000, 15000 and 10000 due to the possibility of the teacher selecting a full training set batchsize and memory constraints.

\subsubsection*{MNIST}
As before, we utilise the MNIST dataset \cite{lecun1998gradient} as another publicly available dataset to assess our results against. We use 30,000 examples for training, 10,000 for validation and 10,000 for the test set. Once again the dataset is partitioned into $N$ datasets to emulate the federated learning approach.

\section{Results}\label{Section:results}
\begin{table*}[t!]\caption{Average classification accuracies and standard deviations for various baseline and state-of-the-art methods on the Ward Admission (tabular), MNIST (image) and CIFAR-10 (image) datasets. All models are averaged over the same five seeds apart from those highlighted with * which indicates that the accuracy reported from the cited text is quoted.}
\vskip 0.15in
\begin{center}
\begin{small}
\begin{sc}
\begin{tabular}{lccccccr}
\toprule
       & Ward Admission & MNIST & CIFAR-10 & CIFAR-10 & CIFAR-10 & eICU\\
       &                &           & Sample 1 & Sample 2 & Sample 3 & \\
Method & Acc (SD) & Acc (SD) & Acc (SD) & Acc (SD) & Acc (SD) & AUC (SD)  \\
\midrule
SMBT    & 0.45 (0.01) & 0.91 (0.01)& 0.65 (0.02)&0.65 (0.02)&0.65 (0.02) & 0.80 (0.02)\\
CURRIC & 0.48 (0.02) & 0.93 (0.02)& 0.68 (0.01)&0.68 (0.01)&0.68 (0.01)&0.81 (0.01) \\ 
DeepFM    & 0.59 (0.01) & N/A &  N/A & N/A & N/A & 0.81 (0.01)    \\
Deep+CrossNet     & 0.58 (0.02) & N/A & N/A & N/A & N/A & 0.82 (0.02)\\
DenseNet*    & N/A & \textbf{0.99} (0.01) & 0.96 (0.01) & 0.96 (0.01) & 0.96 (0.01) & N/A \\
GPipe*      & N/A & \textbf{0.99} (0.01) & \textbf{0.99} (0.01) & \textbf{0.99} (0.01) & \textbf{0.99} (0.01) & N/A \\

RLST      & \textbf{0.62} (0.02) & 0.95 (0.02)& 0.90 (0.01) & 0.90 (0.01) & 0.89 (0.01) & 0.85 (0.02) \\

FLST      & 0.60 (0.02) & 0.94 (0.01)& 0.91 (0.01) & 0.90 (0.01) & 0.91 (0.02) & \textbf{0.86} (0.01) \\
\bottomrule
\end{tabular}
\end{sc}
\end{small}
\end{center}
\vskip -0.1in
\label{table:max_performance_results}
\end{table*}

\begin{table*}[t!]\caption{Average classification accuracies and standard deviations for various baseline and state-of-the-art methods on the hospital ward admission (tabular), eICU (tabular), CIFAR-10 (image) and MNIST (image) datasets. All models are averaged over the same five seeds. We also show how the models perform when subjected to a data poisoning attack and a local model poisoning attack.}
\vskip 0.15in
\begin{center}
\begin{small}
\begin{sc}
\begin{tabular}{lccr}
\toprule
       & Accuracy & Model Poisoning & Data Poisoning \\

Method & Acc (SD) & Acc (SD) & Acc (SD)  \\
\midrule
FedAvg (hospital)    & 0.55 (0.01) & 0.33 (0.01)& 0.47 (0.02)\\
FedMA (hospital) & 0.56 (0.02) & 0.45 (0.02)& 0.52 (0.01) \\ 
FLST (hospital)      & \textbf{0.60} (0.02) & \textbf{0.59} (0.01) & \textbf{0.59} (0.02) \\
FedAvg (CIFAR-10)    & \textbf{0.93} (0.01) & 0.62 (0.01)& 0.65 (0.02)\\
FedMA (CIFAR-10) & 0.93 (0.02) & 0.77 (0.02)& 0.68 (0.01) \\ 
FLST (CIFAR-10)      & 0.90 (0.01) & \textbf{0.85} (0.01)& \textbf{0.87} (0.02)\\
FedAvg (MNIST)    & \textbf{0.95} (0.01) & 0.77 (0.01)& 0.73 (0.02)\\
FedMA (MNIST) & \textbf{0.95} (0.01) & 0.83 (0.01)& 0.81 (0.02) \\ 
FLST (MNIST)      & \textbf{0.95} (0.02) & \textbf{0.86} (0.02) & \textbf{0.88} (0.01) \\
FedAvg - AUC (eICU)    & 0.80 (0.01) & 0.65 (0.07)& 0.69 (0.12)\\
FedMA - AUC (eICU) & 0.82 (0.01) & 0.71 (0.09)& 0.72 (0.04) \\ 
FLST - AUC (eICU)      & \textbf{0.86} (0.02) & \textbf{0.80} (0.04) & \textbf{0.81} (0.03) \\
\bottomrule
\end{tabular}
\end{sc}
\end{small}
\end{center}
\label{table:federated_results}
\end{table*}
As our work lies in the intersection of two research areas within the field of machine learning (namely federated learning and student-teacher learning), we choose to use baselines from both of these fields as comparators. From the student-teacher learning side, we will assess how our method compares in terms of final model performance only. For the federated learning comparison we will compare not only the final model performance but also the robustness of the method to attack.

\subsubsection*{Final Model Performance}
Table \ref{table:max_performance_results} shows how the performance of our federated learning method (FLST) compares to other state-of-the-art classification methods. The baselines we use are the reinforcement learning trained student-teacher setup without scheduling \cite{el2020student} (RLST), two state-of-the-art methods used for classifying tabular data (DeepFM \cite{guo2017deepfm} and Deep+CrossNet \cite{wang2017deep}) and two state-of-the-art classifiers for image recognition (GPipe \cite{huang2019gpipe} and DenseNet \cite{huang2018condensenet}). As baselines we train a standard feedforward neural network (a convolutional neural network for the image datasets) using stochastic mini-batch training (SMBT) and a curriculum (CURRIC) for comparison. 

\begin{figure}[h!]
    \centering
    \includegraphics[scale=0.5]{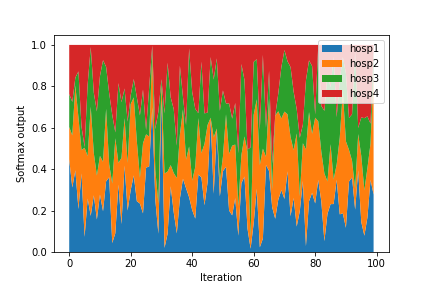}
    \caption{A four-node federated system being scheduled for training a student on the hospital admission location prediction problem. The different colours represent the different nodes.}
    \label{fig:hosp_normal}
\end{figure}

\begin{figure}[h!]
    \centering
    \includegraphics[scale=0.5]{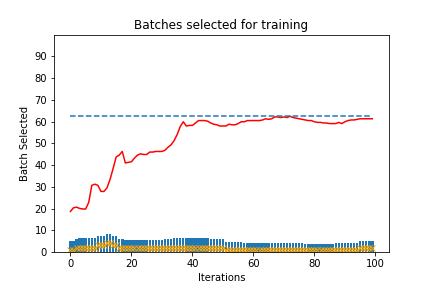}
    \caption{The performance of the student and actions taken by the teachers (scheduled according to Figure \ref{fig:hosp_normal}) at each node to train the student. The orange `x' and blue bar indicate the first and second outputs of the teacher respectively (index of data along the curriculum that is selected and how much data around this to include in that batch). The red line shows the performance of the student on the held-out test set on the hospital admission problem.}
    \label{fig:hosp_normal_student}
\end{figure}

We see that our federated system is capable of producing a performance that is competing with state-of-the-art models that are trained in a centralised manner. Figure \ref{fig:hosp_normal} shows the scheduler outputs during training for the hospital admission problem and Figure \ref{fig:hosp_normal_student} shows the performance of the student and the actions of the teacher during training.

\subsubsection*{Learning from Multiple Hospitals}
When training on the eICU dataset, we utilised data from individual hospitals as the separate nodes in the federated system. A natural question that arises is how the increasing the number of nodes in the system affects the final performance of the student. 

\begin{figure}[h!]
    \centering
    \includegraphics[scale=0.5]{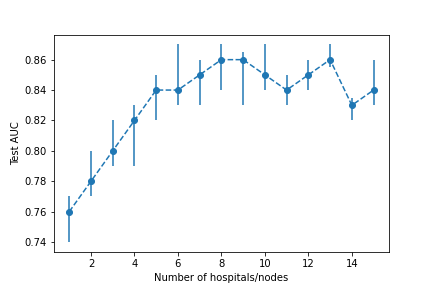}
    \caption{The number of nodes in the federated system (i.e., number of hospitals) versus the student performance at test time. Note that in this figure the seed of the scheduler is held constant and the error bars correspond to the scores generated from using different hospitals for training. So for N=k nodes, the error bars correspond to training only using data from ten different combinations of k hospitals.}
    \label{fig:nodesversusaccuracy}
\end{figure}

In Figure \ref{fig:nodesversusaccuracy} we carry out an ablation on the number of nodes in the system and how it affects performance. In this plot, the scheduler is held at constant seed and the error bars are generated due to selecting different hospitals to train on. There are a total of 208 hospitals in the eICU dataset, but we only select from the top 20 in terms of volume of data recorded. As a result, Figure \ref{fig:nodesversusaccuracy} shows error bars for the difference in performance when ten different combinations of hospital data are used. We can see that with more data being used in the system, the final performance generally increases. We also see that the variance in the performance starts to decrease with the increase in the number of nodes. Given the increase in the volume of data being trained with as nodes are added, this aligns with expectations. The performance seems to plateau indicating that adding more nodes to the system may not necessarily be beneficial for performance. This could be useful for real-world application as it will allow practitioners to prioritise data centres with the highest quality data for their federated systems.

\subsubsection*{Robustness of Federated Training Routine}
Table \ref{table:federated_results} shows how our method performs when compared to other federated learning algorithms. For our baselines we use FedAvg \cite{mcmahan2017communication} where the local models at each node are aggregated before being averaged, as well as FedMA \cite{wang2020federated}, which constructs a shared global model in a layer-wise manner by matching and averaging hidden elements (such as neurons and hidden states). We investigate how performance deteriorates when exposed to different backdoor attacks. We see that our model performs equivalently to state-of-the art federated training setups in terms of test-time performance but outperforms these models when exposed to attack. Through the use of the scheduler, our approach provides an added layer of redundancy in the system thereby allowing attacks to be avoided after their implicit detection through degraded performance on the validation sets stored at all nodes.


\subsubsection*{Iterative Learning}
In our approach we choose to exploit local models to select data and therefore extract gradients that we use to update a global model. The advantage of this approach is that it provides flexibility for a poisoned node within the federated system to be discounted or unused. This can also be done on the fly without the need to inspect the local models after each training run or the data stored at each node. In the following section we look at the ways that a node can be compromised and see how our setup may be able to avoid the global model being poisoned by these scenarios.

\section{Implicit Defensive Setup}
\label{section:defense}
There are various ways in which a federated system can be attacked. In this section, we first show that the scheduler chooses appropriate nodes for training. We then aim to show how using a teaching setup, we can avoid some of the issues that could be faced by a federated learning system under attack.

\subsection{Selecting the Right Teacher}
\begin{figure}[h!]
    \centering
    \includegraphics[scale=0.5]{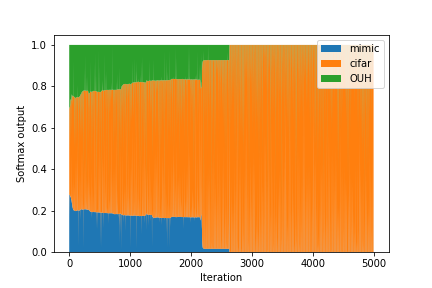}
    \caption{Scheduler selection as the student is trained. The student is being trained on CIFAR-10 and the scheduler learns to use the CIFAR-10 teacher to teach the student.}
    \label{fig:scheduler_CIFAR}
\end{figure}

In our first set of experiments we aimed to see if the scheduler would be able to select the appropriate teacher for the learning task at hand. Our scheduler's task is to select a teacher from three different nodes to train the student. We pre-trained three teachers for separate tasks (hospital admission, CIFAR-10 and MIMIC-III mortality prediction \cite{johnson2016mimic}) and allowed our scheduler to choose from these in order to source the data for training. Figure \ref{fig:scheduler_CIFAR} shows how with training, the scheduler learns to assign the teaching job to the node that contains the teacher trained to teach CIFAR-10 learning. As training progresses, the response from the scheduler becomes entirely dominated by the node in the federated system that corresponds to the appropriate teacher for the task. From this we see that our approach allows for robust teacher selection (further examples of this are included in the supplementary material). 

\subsubsection*{Compromised Data at Nodes}
\begin{figure}[h!]
    \centering
    \includegraphics[scale=0.4]{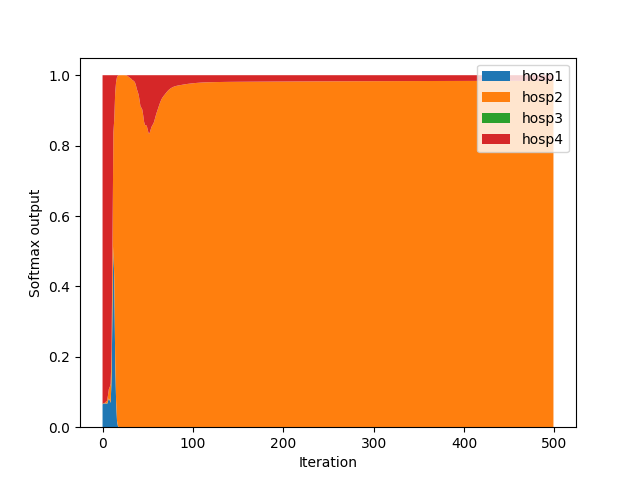}
    \caption{The scheduler selection for the different nodes in the federated system. We see that the scheduler learns to select the only clean dataset for training. The magnitudes of the different colours indicate the softmax output of the scheduler.}
    \label{fig:MNIST_corrupted_nodes}
\end{figure}

\begin{figure}[h!]
    \centering
    \includegraphics[scale=0.4]{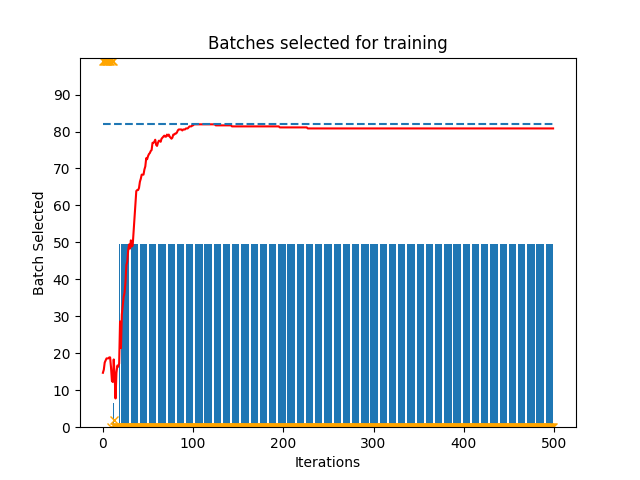}
    \caption{The teacher data selection as well as the performance of the student on the held-out test set for the MNIST digit recognition problem. The orange `x' and blue bars represent the first and second outputs of the teacher respectively. The dashed line shows the maximum accuracy achieved. The red line is the performance of the student on the held-out test set.}
    \label{fig:MNIST_corrupted_student}
\end{figure}

Our next set of experiments investigated the robustness of our method to attack through poisoning of data at local nodes in the federated system. By scheduling training through the use of meta-gradients, we hypothesised that there would be an extra layer of redundancy which would prevent immediate poisoning of the model. In this experimental setup we split our dataset randomly (the size of the splits is also random). We only keep one node clean, with the rest of the datasets on the other nodes being replaced with random values. Figure \ref{fig:MNIST_corrupted_nodes} shows learned scheduling for a student being trained on the MNIST dataset. We see how the scheduler begins with selecting a corrupted dataset before quickly transitioning to selecting another corrupted dataset. The scheduler then selects a dataset with clean data and this selection dominates for the rest of training. In Figure \ref{fig:MNIST_corrupted_student} we see that due to the initial training on corrupted data, the test-set performance degrades. However, as soon as the scheduler learns to use the clean data, the performance improves rapidly. It can also be noticed that the performance achieved is below state-of-the-art. This is likely due to there being a much smaller diversity in trainin data due to node corruption.

\subsubsection*{Compromised Local Models at Nodes}
\begin{figure}[h!]
    \centering
    \includegraphics[scale=0.45]{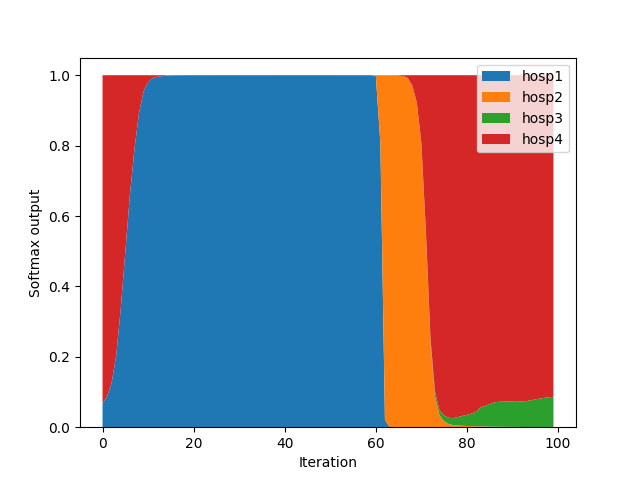}
    \caption{Scheduler selection for a four-node federated system. The hosp1 node (orange) corresponds to the poisoned teacher (randomised weights). We see that after brief selection, the scheduler reduces the contribution of this node. The student is being trained on the MNIST task.}
    \label{fig:corrupt_teacher}
\end{figure}

The next experiment we investigate is how our system trains when there are compromised local models (whether it is the student or the teacher). We do this by replacing one of the weights of the teacher on one of the local nodes with random values. Due to the setup, the effect of having a compromised student or teacher is the same: a high loss which encourages the scheduler to change its selection. Figure \ref{fig:corrupt_teacher} shows how the scheduler selects nodes to train from. We see initially nodes 1 and 0 are used to train before the scheduler attempts to use the poisoned teacher. After repeated reductions in the federated validation sets, the scheduler rapidly changes its selection favouring nodes 4 and increasingly 3. With further training, we see this rapid removal of training using node 1 continue whenever it is encountered.

\subsubsection*{Transfer Learning}
Due to the use of a scheduler, there is potential for exploiting transfer learning in order to train students on tasks that there are no trained teachers for. We hypothesise that through exploiting the various teacher's skills in succession, we can provide the gradients that the student needs in order to master the unseen task. Experiments to test this hypothesis are included in the supplementary material.

\section{Discussion and Conclusion}
In this work we have shown that using a federated system, with teachers at the nodes that select the data to provide gradient updates to a central model, can achieve state-of-the-art performance as well as protect the privacy of patients. We have further shown that the setup provides some protection against attacks on data stored at each node or the local models being used at each node. 
\par However, there remain some challenges associated with this approach that need to be addressed in order for it to become practicable. The first is that the training of the centralised model (student) is inherently unstable due to the continual training that this setup expects. It is expected that the student performance will converge with training, however should data poisoning occur, the scheduler will need to continue a few iterations of training in order to recover a well-performing model. However, posing the problem in this way also provides flexibility for growing datasets at each node. All that would need to happen would be the re-sorting of the curriculum at each node and the scheduler and teachers could be used as before.
\par Another limitation is the need for centralised control. It is important in this setup that all the hospitals communicate their responses to the central node for actions to be taken by the scheduler. In the case of large institutions such as hospitals, this may be acceptable, but is unlikely to be for faster-paced learning environments such as learning from mobile phones, where interruptions to communication can be frequent. However, upon re-connection to the federated system, any reductions in performance to the whole system will be used as signals to improve the scheduler selection and with training the performance should recover.
\par Furthermore, in order to ensure diversity in selections by the scheduler we introduced the entropy loss that discouraged convergence on one selection. This may be a naive way of encouraging diversity in selection and we believe that there may be better additive losses and regularisation terms that can be used to design a loss function that will serve the purposes of the scheduler better.
\par For further protection against attack, sentry agents (much like the teachers) could also be trained to detect any anomalies or designed attacks within the batches selected by the teachers before the losses are passed onto the central node. This would reduce the burden of scanning the entire dataset at the node before training.
\par To conclude, we believe we have presented a promising direction for federated learning between large institutions such as hospitals. With further work, we believe that we can develop this into a robust system that can continually learn from growing datasets while maintaining a state-of-the-art performance for the task at hand. 

\newpage

\newpage
\bibliography{Bibliography.bib}
\bibliographystyle{icml2020}

\appendix
\section{Features}

\begin{table}[h!]
\centering
\begin{tabular}{@{}|l|l|@{}}
\hline
Bacteriology test requested? & Biochemical tests requested? \\ \hline Blood cultures requested & Blood gas test requested? \\ \hline
CT scan requested? & Cardiac enzyme test requested? \\ \hline
Clotting study requested & Blood cross-matching requested? \\ \hline
 Heart rate at entry & Dental investigation requested?  \\ \hline
ECG requested? & Haematology test requested? \\ \hline
MRI scan requested?  & Immunology test requested?  \\ \hline
Continuous vitals requested? & No tests requested \\ \hline
Orthopedic tests requested? & Other tests requested?  \\ \hline
Pregnancy test requested? & Respiratory rate at entry  \\ \hline
Serology test requested? & Previous specialty \\ \hline
Toxicology test requested? & Ultrasound test requested? \\ \hline
Urine test requested? & X-ray scan requested? \\ \hline
Admission method & Admission source  \\ \hline
Age & Experiencing atrial fibrillation?  \\ \hline
\# historic diagnoses & Previous management \\ \hline
Previous admission to ED? &  Ethnic category  \\ \hline
Frequently admitted? & Gender  \\ \hline
Dist. of address to hospital & No. Investigations requested \\ \hline
Previous ED visit days ago & Previous visit LOS  \\ \hline
Mortality indicator severity score & Historic diagnosis codes \\ \hline
\end{tabular}
\caption{Table containing the patient specific features available at initial medical assessment that were used in the OUH dataset.}
\label{table: features table}
\end{table}

\begin{table}[h!]
\centering
\begin{tabular}{@{}|l|l|@{}}
\hline
Hours of sunlight & Hour admitted to ED  \\ \hline
Month admitted to ED & \# ED Attendees in last 12 hours \\ \hline
\# ED Attendees in last 4 hours & \# ED Attendees in last 8 hours \\ \hline
\# ED Attendees in last hour & \# Breaches last 12 hours \\ \hline
\# Breaches last 4 hours &\#  Breaches last 8 hours \\ \hline
\# Breaches last hour &  Has it rained today? \\ \hline
Min. day temp (degrees) & Max. day temp (degrees) \\ \hline
Weekday (one-hot) &  \\ \hline

\end{tabular}
\caption{Table containing the environmental/hospital features that were used in the OUH dataset.} 
\label{table: features LASSO table}
\end{table}

\begin{table}[h!]
\centering
\begin{tabular}{@{}|l|l|@{}}
\hline
Heart rate & Mean arterial pressure \\ \hline
Diastolic blood pressure & Systolic blood pressure \\ \hline
Blood oxygen saturation & Respiratory rate  \\ \hline
Temperature & Glucose \\ \hline
Oxygen delivery rate & pH \\ \hline
Height & Weight\\ \hline
Age & Admission diagnosis \\ \hline
Ethnicity & Gender \\ \hline
Glasgow coma score total & \\ \hline

\end{tabular}
\caption{Table containing the features used for mortality prediction on the eICU dataset. Features are extracted from the first measurement of each patient upon admission to ICU.} 
\label{table:eICU features}
\end{table}

\end{document}